\pdfoutput=1
\documentclass[11pt]{article}

\usepackage[dvipsnames,table]{xcolor}
\usepackage{multibib}
\usepackage{graphicx}
\usepackage{tabularx}
\usepackage{soul}
\usepackage{subfig}
\usepackage{multirow}
\usepackage{arydshln}
\usepackage{enumitem}
\usepackage{booktabs}

\usepackage{hyperref}
\usepackage{xstring}

\usepackage{color}

\newcommand{\efrEnglish}{\texttt{MELD-FR}}
\newcommand{\ercCM}{\textsc{E-MaSaC}}
\newcommand{\efrCM}{\textsc{EFR-MaSaC}}
\usepackage[final]{acl}
\usepackage{times}
\usepackage{latexsym}
\usepackage[T1]{fontenc}
\usepackage[utf8]{inputenc}
\usepackage{microtype}
\usepackage{inconsolata}

\title{SemEval 2024 -- Task 10: Emotion Discovery and Reasoning its Flip in Conversation (EDiReF)}

\author{Shivani Kumar$^1$, Md Shad Akhtar$^1$, Erik Cambria$^2$, Tanmoy Chakraborty$^3$ \\
  $^1$ IIIT Delhi, India; $^2$ NTU, Singpore; $^3$ IIT Delhi, India \\
  \texttt{\{shivaniku, shad.akhtar\}@iiitd.ac.in,camria@ntu.edu.sg } \\ \texttt{tanchak@iitd.ac.in} }

\begin{document}
\maketitle
\begin{abstract}
We present SemEval-2024 Task 10, a shared task centred on identifying emotions and finding the rationale behind their flips within monolingual English and Hindi-English code-mixed dialogues. This task comprises three distinct subtasks -- emotion recognition in conversation for code-mixed dialogues, emotion flip reasoning for code-mixed dialogues, and emotion flip reasoning for English dialogues. Participating systems were tasked to automatically execute one or more of these subtasks. The datasets for these tasks comprise manually annotated conversations focusing on emotions and triggers for emotion shifts.\footnote{The task data is available at \url{https://github.com/LCS2-IIITD/EDiReF-SemEval2024.git}.} A total of $84$ participants engaged in this task, with the most adept systems attaining F1-scores of $0.70$, $0.79$, and $0.76$ for the respective subtasks. This paper summarises the results and findings from $24$ teams alongside their system descriptions.
\end{abstract}

\section{Introduction}
\label{sec:intro}
In pursuit of one of AI's ultimate objectives, i.e., emulating human behaviour, machines must comprehend human emotions \cite{ekman,Picard:1997:AC:265013}. Consequently, Emotion Recognition in Conversation (ERC) has emerged as a vibrant domain within NLP \cite{hazarika-etal-2018-conversational,hazarika-etal-2018-icon,zhong-etal-2019-knowledge,ghosal-etal-2019-dialoguegcn,aghmn}. The significance of emotion detection amplifies particularly during shifts in the speaker's emotional state. However, merely identifying an emotional transition is insufficient; understanding the catalyst behind the shift is crucial for facilitating informed decisions by other speakers. For instance, identifying the utterance responsible for a customer's transition from a positive emotional state (e.g., joy) to a negative one (e.g., disgust) due to a flawed dialogue system is critical in customer service. Such insights can serve as feedback to the dialogue system, enabling it to avoid (negative emotion-flip) or replicate (positive emotion-flip) similar utterances to enhance the customer experience in the future.

\begin{figure}[t]
    \centering
    \subfloat[Emotion-flip is caused by more than one utterance.\label{fig:trigger:example:multiple}]{
    \includegraphics[width=\columnwidth]{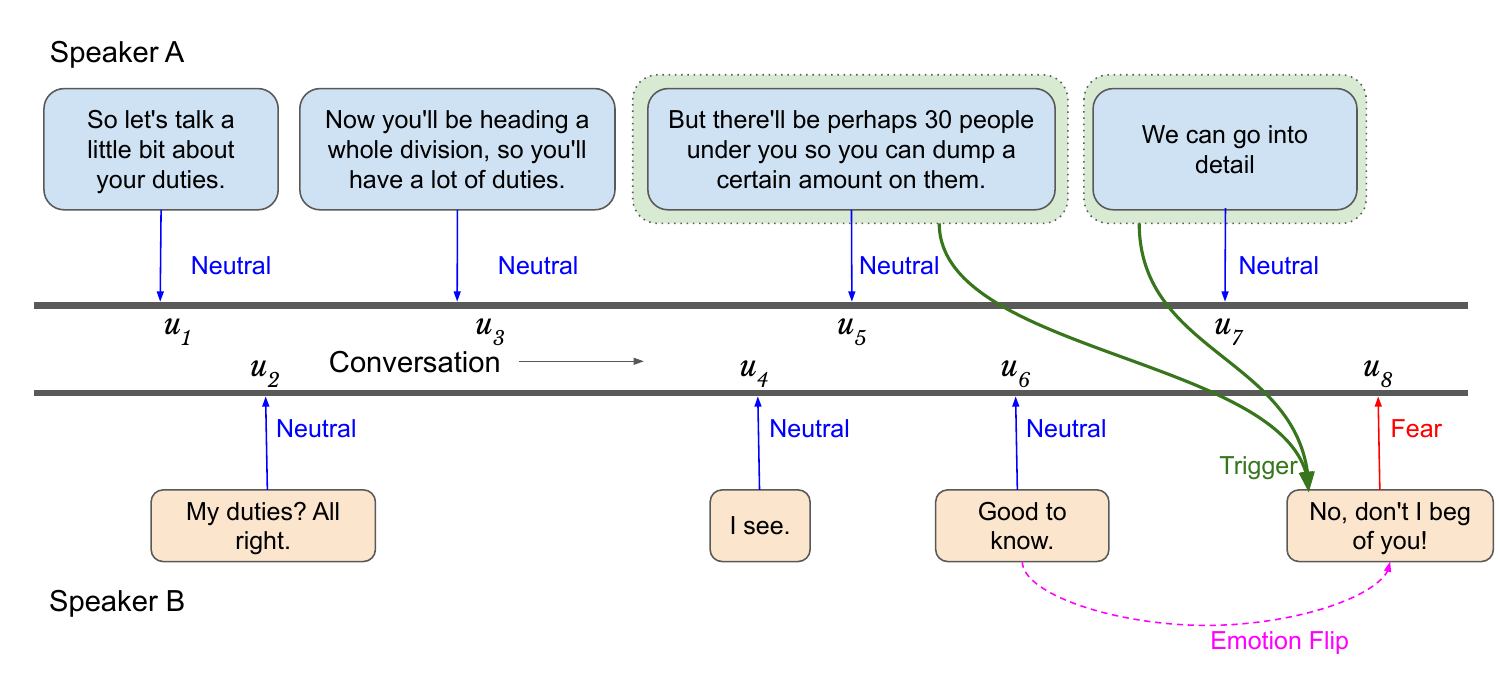}}\\
    \subfloat[Emotion-flip is caused by the previous utterance. Out of five emotion-flips, we show only two of them ($u_5\rightarrow u_7$ and $u_6 \rightarrow u_8$) for brevity. Other emotion-flips are $u_1\rightarrow u_3$, $u_2\rightarrow u_4$, and $u_4\rightarrow u_6$ with triggers $u_3$, $u_3$, and $u_6$, respectively.\label{fig:trigger:example:normal}]{
    \includegraphics[width=\columnwidth]{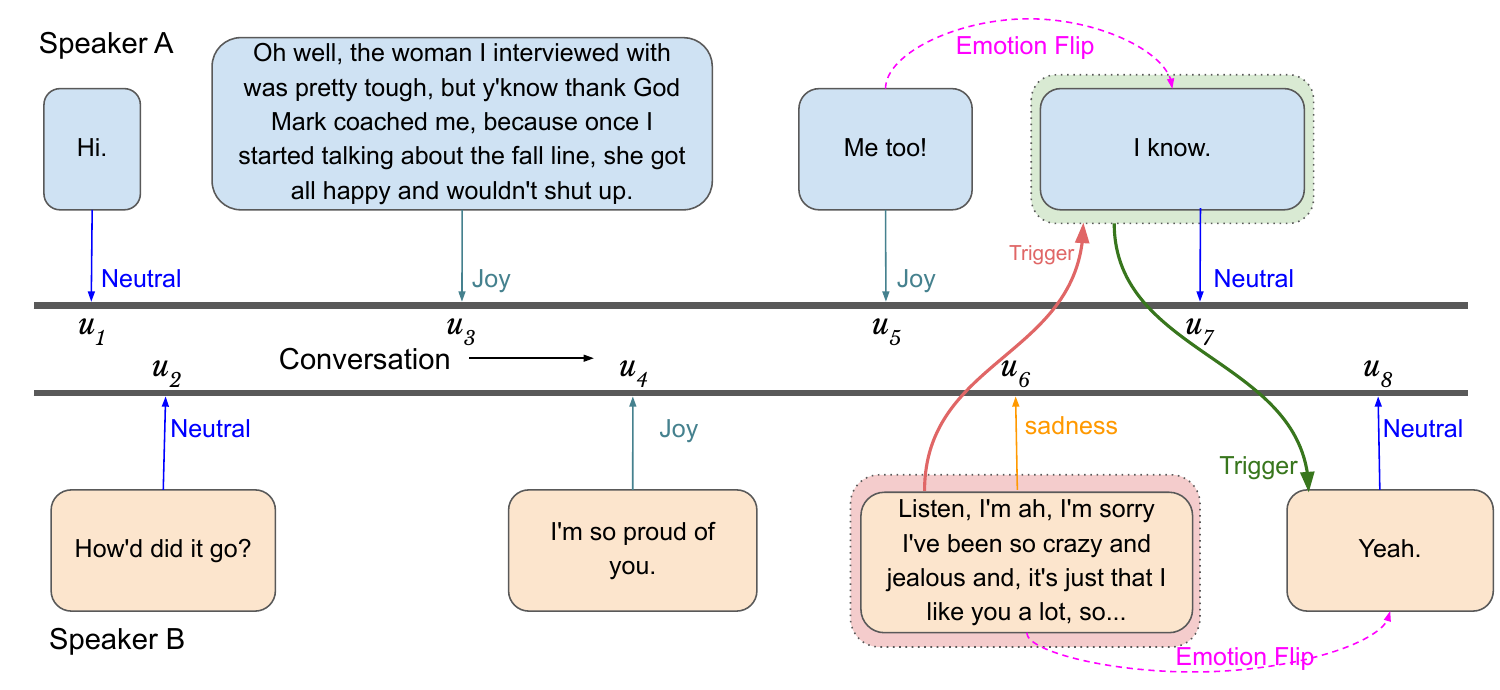}}
    \caption{Examples of emotion-flip reasoning.}
    \label{fig:efr:examples}
    \vspace{-5mm}
\end{figure}

Emotion-Flip Reasoning (EFR), as outlined by \citet{kumar2021discovering}, presents a novel endeavour aimed at pinpointing the trigger utterances responsible for an emotion-flip within the context of a multi-party conversation. Figure \ref{fig:efr:examples} provides illustrative scenarios depicting the essence of EFR. In Figure \ref{fig:trigger:example:multiple}, Speaker B undergoes a transition in emotion (\textit{neutral}$\rightarrow$\textit{fear}) between utterances $u_6$ and $u_8$. Notably, this emotional shift can be attributed to the contributions of Speaker A through utterances $u_5$ and $u_7$. The fundamental objective of this task is to discern these trigger utterances ($u_5$ and $u_7$) given a target emotion-flip utterance ($u_8$) and the context preceding it ($u_1 \dots u_7$).

As an effort to advance research within the domain of ERC and EFR, this shared task seeks to assess the efficacy of NLP systems in automatically addressing both of these tasks. Furthermore, as technological applications extend beyond English to encompass non-English, multilingual, and code-mixed populations, there is a growing need to broaden the scope of research. To support this cause and further the exploration of code-mixed languages, we advocate for the inclusion of ERC and EFR tasks within Hindi-English (Hinglish) code-mixed conversations. Specifically, the shared task is segmented into three distinct subtasks:
\begin{itemize}[leftmargin=*]
\itemsep0em
    \item \textbf{\textit{Task A --} ERC in Hindi-English code-mixed conversation:} Given a multiparty code-mixed conversation, tag each utterance with one of the eight emotion labels -- \textit{anger}, \textit{disgust}, \textit{fear}, \textit{sadness}, \textit{surprise}, \textit{joy}, \textit{contempt}, and \textit{neutral}.
    \item \textbf{\textit{Task B --} EFR in Hindi-English code-mixed conversation:} Given a multiparty code-mixed conversation along with emotions for each utterance, the goal is to identify the trigger utterance for each emotion-flip in the dialogue.
    \item \textbf{\textit{Task C --} EFR in English conversation:} It is similar to Task B but in monolingual English.
\end{itemize}

The decision to omit ERC for monolingual English stems from its thorough examination and the abundance of available datasets. Conversely, ERC in Hindi-English code-mixed conversation remains relatively unexplored, and as far as we are aware, no other dataset besides the one outlined in this article is publicly accessible.

Further elaboration on our task data and setting is provided in Sections \ref{sec:data} and Section \ref{sec:task}, respectively. The participating teams are outlined in Section \ref{sec:participants}, with their task outcomes and assessments detailed in Section \ref{sec:results}.

\section{Related Work}
\label{sec:relatedwork}
\paragraph{Emotion recognition.} Identifying emotions has been a focal point in prior research, with investigations into emotion analysis \cite{ekman,Picard:1997:AC:265013,cowen2017self,mencattini2014speech,zhang2016intelligent,cui2020eeg} initially centring on standalone inputs devoid of contextual cues. However, recognising the significance of contextual information, the emphasis shifted towards emotion detection within conversations, particularly ERC. Initially, ERC was tackled using heuristic approaches and conventional machine learning techniques \cite{fitrianie2003multi, chuang2004multi, li2007research}. However, the recent trend has witnessed a transition towards the adoption of a diverse array of deep learning methodologies \cite{hazarika-etal-2018-icon, zhong-etal-2019-knowledge, bieru, ghosal-etal-2019-dialoguegcn, aghmn, tlerc, dialogxl, poria2017context, jiao2019higru, 9706271, yang2022hybrid, ma2022multi}.

\paragraph{Emotion and code-mixing.} Current studies addressing emotion analysis in code-mixed language primarily centre around isolated social media texts \cite{SASIDHAR20201346, 10.1145/3552515, wadhawan2021emotion} and reviews \cite{Suciati2020, ZHU2022107436}. Despite examinations into aspects like sarcasm \cite{kumar-etal-2022-become, kumar2022explaining}, humour \cite{10.1109/TAFFC.2021.3083522}, and offence \cite{madhu2023detecting} within code-mixed conversations, the domain of emotion analysis remains largely uncharted, lacking pertinent literature, to the best of our knowledge. Our objective is to address this gap by delving into the under-explored realm of ERC, specifically within Hindi-English code-mixed dialogues in this shared task.

\paragraph{Beyond emotion recognition.} The interpretability of emotion recognition within the linguistic domain represents a relatively uncharted avenue of research, with only a limited number of studies delving into this field. Previous works by \citet{lee-etal-2010-text, poria2021recognizing, 9969873} have focused on investigating the root causes of expressed emotions in text, commonly referred to as 'emotion-cause analysis.' This task involves identifying a specific span within the text that elicits a particular emotion. While on an abstract level, both emotion-cause analysis and emotion-flip reasoning tasks may appear interconnected, they diverge significantly in practice. Emotion-cause analysis aims to pinpoint phrases within the text that provide clues or triggers for the expressed emotion. In contrast, our proposed EFR task pertains to conversational dialogues involving multiple speakers, with the objective of extracting the causes \cite{10164178} or triggers behind emotional transitions for a speaker. The triggers comprise one or more utterances from the dialogue history, as illustrated in the two examples in Figure \ref{fig:efr:examples}.

\begin{table*}[t]
\centering
    \subfloat[ERC -- English]{
    \resizebox{0.6\textwidth}{!}
    {
    \begin{tabular}{c|c|c|c|c|c|c|c|c}
     \multirow{2}{*}{\bf Split} & \multicolumn{7}{c|}{\bf Emotions} & \multirow{2}{*}{\bf Total} \\ \cline{2-8}
     & \bf Disgust & \bf Joy & \bf Surprise & \bf Anger & \bf Fear & \bf Neutral & \bf Sadness & \\ \hline \hline
         \bf Train  & 225 & 1466 & 1021 & 911 & 229 & 3702 & 576 & 8130 \\
         \bf Dev & 20 & 156 & 144 & 126 & 39 & 395 & 97 & 977 \\
         \bf Test & 61 & 325 & 238 & 283 & 42 & 943 & 169 & 2061 \\ \hline
    \end{tabular}}}
    \subfloat[EFR -- English]{
\resizebox{0.37\textwidth}{!}
    {
\begin{tabular}{c|c|c|c}
\\
         \bf Split & \bf \#D with Flip & \bf \#U with Flip & \bf \#Triggers  \\ \hline \hline 
         \bf Train & 834 & 4001 & 6740 \\
         \bf Dev & 95 & 427 & 495 \\
         \bf Test & 232 & 1002 &  1152  \\ \hline
    \end{tabular}}}
    
       \subfloat[ERC -- Hindi]{
    \resizebox{0.63\textwidth}{!}
    {
    \begin{tabular}{c|c|c|c|c|c|c|c|c|c}
     \multirow{2}{*}{\bf Split} & \multicolumn{8}{c|}{\bf Emotions} & \multirow{2}{*}{\bf Total} \\ \cline{2-9}
     & \bf Disgust & \bf Joy & \bf Surprise & \bf Anger & \bf Fear & \bf Neutral & \bf Sadness & \bf Contempt & \\ \hline \hline
         \bf Train  &  127 & 1646 & 444  & 856 & 530 & 4091 & 572 & 549 & 8815 \\
         \bf Dev & 21 & 242 & 68 & 122 & 91 & 652 & 132 & 75 & 1403 \\
         \bf Test & 21 & 382 & 57 & 150 & 129 & 697 & 167 & 87 & 1690 \\ \hline
    \end{tabular}}}
    \subfloat[EFR -- Hindi]{
\resizebox{0.35\textwidth}{!}
    {
\begin{tabular}{c|c|c|c}
\\
         \bf Split & \bf \#D with Flip & \bf \#U with Flip & \bf \#Triggers  \\ \hline \hline 
         \bf Train & 344 & 4406 & 5565 \\
         \bf Dev & 47 & 686 & 959 \\
         \bf Test & 58 & 781 & 1026 \\ \hline
    \end{tabular}}}
    
    \caption{Statistics of the English and Hindi datasets for ERC and EFR.}
    \label{tab:dataset}
    \vspace{-5mm}
\end{table*}

In this shared task, we tackle the challenge of automatically performing the task of ERC and EFR for code-mixed and monolingual English dialogues in order to further this research direction.

\section{Data}
\label{sec:data}
\paragraph{English Conversations:}
We extend MELD \cite{poria-etal-2019-meld}, an established ERC dataset comprising monolingual English dialogues, by incorporating annotations for emotion-flip reasoning. These dialogues are sourced from the popular TV series \textit{F.R.I.E.N.D.S}\footnote{\url{https://www.imdb.com/title/tt0108778/}}. Each utterance $u$ is attributed to a specific speaker $s$ and assigned an emotion label $e \in $[\textit{anger}, \textit{disgust}, \textit{fear}, \textit{sadness}, \textit{surprise}, \textit{joy}, \textit{neutral}]. In the context of a speaker's emotional transition, we designate and label trigger utterances as $1$ if they induce the speaker's emotional shift -- the emotion alters from the speaker's preceding utterance within the same dialogue. In contrast, a label $0$ indicates that the utterance bears no responsibility for the emotional transition.

To facilitate the annotation of triggers, we establish a set of guidelines outlined below. Within this framework, a \textbf{\textit{trigger}} is defined as any utterance within the contextual history of the target utterance (the utterance for which the trigger is to be identified) meeting the following criteria:
\begin{enumerate}[leftmargin=*]
\itemsep0em
    \item An utterance, or part thereof, directly influencing a change in emotion of the target speaker is designated as the trigger.
    \item The speaker of the trigger utterance may be different from or the same as the target speaker.
    \item The target utterance itself may qualify as a trigger utterance if it contributes to the emotional transition of the target speaker. For instance, if an individual's emotion shifts from \textit{neutral} to \textit{sad} due to conveying a sad message, then the target utterance is deemed responsible for the transition.
    \item Multiple triggers may be accountable for a single emotional transition.
    \item In cases where the rationale behind an emotional transition is not identifiable from the data, no utterance should be labelled as a trigger.
\end{enumerate}

In total, we have annotated emotion-flip reasoning for $1,161$ monolingual English conversation dialogues, encompassing $8,387$ trigger utterances across $5,430$ emotion-flip instances. Three annotators carefully annotated these dialogues in accordance with the aforementioned guidelines for trigger identification. Among the three, two annotators were male while one was female, all possessing 3-10 years of research experience within the 30-40 age bracket. We calculated the alpha-reliability inter-annotator agreement \cite{krippendorff2011computing} between each pair of annotators, yielding $\alpha_{AB}=0.824$, $\alpha_{AC}=0.804$, and $\alpha_{BC}=0.820$. By averaging these scores, we derived an overall agreement score of $\alpha=0.816$. We call the resultant dataset as \efrEnglish.

\paragraph{Hindi-English Code-mixed Conversations:}
For code-mixed tasks, we adhere to identical guidelines as those applied to English, selecting code-mixed conversations from a preexisting dialogue dataset called MaSaC \cite{10.1109/TAFFC.2021.3083522}. The dialogues in the dataset are sourced from the popular Indian TV series `\textit{Sarabhai vs Sarabhai}'\footnote{\url{https://www.imdb.com/title/tt1518542/}}. Further, we annotated $11,908$ utterances spanning $449$ dialogues, encompassing eight emotion labels (including `\textit{contempt}' alongside the six basic emotions and \textit{neutral}) for the ERC task, achieving a Krippendorff alpha-reliability inter-annotator agreement \cite{krippendorff2011computing} of $0.85$. In the context of EFR, we annotated $7,550$ trigger utterances for $5,873$ emotion-flip occurrences. Mirroring our approach with the English dataset, we engaged experts fluent in both Hindi and English to ensure accuracy. As a measure of quality assurance, the Krippendorff alpha-reliability inter-annotator agreement stands at $\alpha = 0.853$. The resultant dataset is denoted as \ercCM\ and \efrCM\ for the ERC and EFR tasks, respectively. A concise overview of both datasets is presented in Table \ref{tab:dataset}.

\section{Task and Background}
\label{sec:task}
The idea of the presented shared task is to delve into ERC and EFR within the domain of English and code-mixed dialogues. This section delves into our preliminary investigations for the three subtasks entailed in this collaborative endeavour.

\subsection{Shared Task Settings}

\textbf{Task A.} In the task of ERC within code-mixed dialogues, participants receive textual utterances as input along with their respective speakers for each dialogue. Their objective is to develop systems capable of autonomously predicting the emotion labels for each utterance. Essentially, the system is presented with a dialogue $D_{erc} = \{(s_1, u_1), (s_2, u_2), ..., (s_n, u_n)\}$, and it must anticipate the emotions $e_i$ for each utterance $u_i$ uttered by speaker $s_i$. Weighted F-1 score of the emotion classification is used as the evaluation metric for the task of ERC.\\
\textbf{Task B.} For the code-mixed EFR task, participants receive dialogues along with their corresponding utterances, speakers, and emotions, presented in the format $D_{efr} = \{(s_1, u_1, e_1), (s_2, u_2, e_2), ..., (s_n, u_n, e_n)\}$. Their objective is to anticipate trigger utterances, $T$, from the context whenever a speaker undergoes an emotion flip. In other words, $T \in \{u_i, ..., u_j\}$ if $s_i = s_j$ and $e_i \neq e_j$. The evaluation metric of choice for this task is the F1 score obtained for trigger utterances.\\ 
\textbf{Task C.} The input modelling for Task C mirrors that of Task B, as both tasks revolve around EFR. Here, the data comes from the \efrEnglish\ dataset and is present in Monolingual English. Just like Task B, the evaluation is conducted based on the F1 score achieved for trigger utterances.

\subsection{Pilot Study}
\textbf{Task A.}
Our preliminary investigation for the task of ERC in code-mixed setting \cite{kumar-etal-2023-multilingual}, we integrate commonsense knowledge with the dialogue representation acquired from a backbone architecture designed for dialogue understanding. We leverage the COMET graph \cite{bosselut-etal-2019-comet} to extract commonsense knowledge, and subsequently employ context-aware attention \cite{Yang_Li_Wong_Chao_Wang_Tu_2019} to integrate this information with the dialogue context. This adaptable module, when combined with RoBERTa \cite{liu2019roberta}, yields a weighted average F1-score of $0.44$ in performance.\\
\textbf{Task B.}
For evaluating the feasibility of our second subtask, we employ FastText multilingual word embeddings\footnote{\url{https://fasttext.cc/docs/en/crawl-vectors.html}} for the tokens and perform classification using the proposed model for Task C to obtain an F1-score of $0.27$ for trigger identification.\\
\textbf{Task C.}
In our initial exploration \cite{kumar2021discovering}, we explored a memory-network and transformer-based architecture to address each occurrence of emotion-flip. This approach yielded a trigger-F1 score of $0.53$. While these findings surpassed various baselines, the overall performance remains inadequate from a practical standpoint, with an error rate of approximately $50\%$.

\section{Participants}
\label{sec:participants}
A total of $84$ participants engaged in the CodaLab competition organised for the shared task\footnote{\url{https://codalab.lisn.upsaclay.fr/competitions/16769}}, with $24$ teams submitting papers describing their systems. Among the submissions, a prevailing trend emerges with the widespread adoption of Large Language Models (LLMs) such as BERT \cite{devlin2018bert}, RoBERTa \cite{liu2019roberta}, GPT \cite{radford2019language}, LLaMa \cite{touvron2023llama}, and Mistral \cite{jiang2023mistral}. Techniques such as fine-tuning, instruction tuning, ensembling, and prompting significantly contribute to enhanced performance in the task. Moreover, many approaches utilise machine learning-based methods including linear regression and SVM. Additionally, some studies explore statistical and rule-based methods such as TF-IDF. While LLMs dominate the approaches for both ERC and EFR, machine learning methods also remain popular among the participants. Furthermore, there appears to be a notable preference among teams for Task A over Tasks B and C, as evidenced by the higher participation in Task A compared to the latter two. An overview of the top-performing models from various teams for ERC is provided in Table \ref{tab:papersERC}, while Table \ref{tab:papersEFR} presents the systems for EFR. We summarize some of the techniques used by the top performing systems below.

\paragraph{Using LLMs} There exists a prevalent preference for LLMs among teams addressing the ERC and EFR tasks, with approximately $18$ methods leveraging LLMs for these endeavours. Notably, BERT and its variants emerge as the most favoured models. Some teams explore larger open-source language models like Zephyr \cite{tunstall2023zephyr} and Mistral, while at least one team delves into closed-source alternatives such as GPT3.5 \cite{NEURIPS2020_1457c0d6}. In the realm of ERC, the leading system (refer to Table \ref{tab:erc_results}) integrates DistilBERT \cite{sanh2020distilbert} with classical machine learning techniques to execute emotion classification optimally. Although the authors experiment with BERT, RoBERTa, and GPT-4 \cite{openai2023gpt4}, their most effective model combines DistilBERT with classical ML algorithms. They adopt a two-step approach, initially extracting contextual features from dialogues using an LLM, then inputting these features into classical ML algorithms such as random forests, SVM, logistic regression, and Naive Bayes. Notably, DistilBERT outperforms GPT-4, possibly attributed to the latter's extensive parameter count, necessitating substantial data for meaningful learning. However, our Task A dataset (\ercCM) encompasses only $\sim8500$ utterances, limiting the efficacy of larger models. Conversely, lighter models like DistilBERT exhibit superior adaptability with limited data, capturing nuanced patterns effectively. This finding aligns with observations from various teams, including BITS Pilani, where BERT outperforms Llama.

For the task of EFR as well, LLMs appear to be the predominant choice among the teams. However, intriguingly, the most effective model for this task (refer to Table \ref{tab:efr_cm_results}) adopts a classical machine learning approach - XGBoost. Further elaboration on this aspect is provided in Section \ref{sec:efr_analysis}.

\begin{table}[t]
\centering
\resizebox{\columnwidth}{!}{%
\begin{tabular}{l|l|l}
\hline
\textbf{Team Name} & \textbf{Backbone Architecture} & \textbf{Model Type} \\ \hline
AIMA               & GPT3.5 + ML                    & Ensemble            \\ %\hline
BITS Pilani        & BERT                           & LLM                 \\ %\hline
CLTeam1            & RoBERTa \& BERT                & LLM+Ensemble        \\ %\hline
FeedForward        & Zephyr                         & LLM                 \\ %\hline
Hidetsune          & SpaCy-v3                       & ML                  \\ %\hline
IASBS              & DistilBERT + ML                & LLM+ML              \\ %\hline
IITK               & Transformer + GRU              & LLM+DL              \\ %\hline
INGEOTEC           & Bag of Words                   & Statistical         \\ %\hline
Innovators         & SVM                            & ML                  \\ %\hline
ISDS-NLP           & RoBERTa                        & LLM                 \\ %\hline
MorphingMinds      & LR                             & ML                  \\ %\hline
RACAI              & BERT + ML                      & LLM+ML              \\ %\hline
SSN\_ARMM          & TF-IDF                         & Statistical         \\ %\hline
SSN\_Semeval10     & BERT                           & LLM                 \\ %\hline
TECHSSN            & LSTM                           & DL                  \\ %\hline
TECHSSN1           & RoBERTa                        & LLM                 \\ %\hline
TransMistral       & Mistral 7B                     & LLM                 \\ %\hline
TW-NLP             & MBERT                          & LLM                 \\ %\hline
UCSC NLP           & BERT                           & LLM                 \\ %\hline
UMUTeam            & BERT                           & LLM                 \\ %\hline
VerbaNexAI Lab     & Transformer + GRU              & LLM+DL              \\ %\hline
YNU-HPCC           & DeBERTa                        & LLM                 \\ \hline
\end{tabular}%
}
\caption{Summary of the models according to the submitted system descriptions for Task A (ERC).}
\label{tab:papersERC}
\end{table}

\paragraph{Classical machine learning and deep learning methods} Efficiently capturing context information is crucial in modelling conversations. Several teams explored this aspect, utilising Recurrent Neural Networks (RNNs) like LSTMs and GRUs. Specifically, at least three teams have integrated GRU with Transformers to enhance context capture. Conversely, team TECHSSN adopts a simpler approach, employing LSTM with intelligent embedding layers for both ERC and EFR tasks. However, these methods frequently fall short in comparison to utilising pre-trained LLMs, as outlined in Section \ref{sec:efr_analysis}.

\begin{table}[t]
\centering
\resizebox{\columnwidth}{!}{%
\begin{tabular}{l|l|l}
\hline
\textbf{Team Name} & \textbf{Backbone Architecture} & \textbf{Model Type} \\ \hline
FeedForward        & Zephyr                         & LLM                 \\ %\hline
GAVx               & GPT3.5                         & LLM                 \\ %\hline
IASBS              & DistilBERT + ML                & LLM+ML              \\ %\hline
IITK               & Transformer + GRU              & LLM+DL              \\ %\hline
Innovators         & -                              & Rule Based          \\ %\hline
LinguisTech        & -                              & NER Model           \\ %\hline
SSN\_ARMM          & TF-IDF                         & Statistical         \\ %\hline
TECHSSN            & LSTM                           & DL                  \\ %\hline
TW-NLP             & XGBoost                        & ML                  \\ %\hline
UCSC NLP           & BERT + GRU                     & LLM+DL              \\ %\hline
UMUTeam            & BERT                           & LLM                 \\ %\hline
YNU-HPCC           & DeBERTa                        & LLM                 \\ \hline
\end{tabular}%
}
\caption{Summary of the models according to the submitted system descriptions for Task B and C (EFR).}
\label{tab:papersEFR}
\end{table}

\paragraph{Rule-based and statistical methods} The surge in deep learning's popularity can be attributed to the remarkable advancements in LLMs. This has led to a decline in the usage of traditional rule-based or statistical approaches, despite their potential to perform comparably in certain scenarios alongside more intricate machine learning or deep learning methods. It was pleasant to observe numerous teams incorporating such traditional techniques into this shared task. Notably, at least four teams opted for methods like Bag of Words, TF-IDF, NER based, and rule based approaches. While these methods may not excel in the ERC task, they surprisingly demonstrate superiority in the EFR task. This is reasoned in detail in Section \ref{sec:efr_analysis}.

\section{Results}
\label{sec:results}
In this section, we delve into the outcomes achieved by the participating teams in the shared task outlined earlier. Initially, we will examine the results submitted by the $24$ teams, which provided detailed descriptions of their systems. Subsequently, we will present the leaderboard, showcasing the performance rankings of all participants.

\begin{table}[t]
\centering
\resizebox{\columnwidth}{!}{%
\begin{tabular}{l|p{15em}|l}
\hline
\textbf{Rank} & \textbf{Team Name} & \textbf{Results} \\ \hline
3                     & IASBS              & 0.70             \\
5                     & FeedForward        & 0.51             \\
6                     & TW-NLP             & 0.46             \\
7                     & TECHSSN1           & 0.45             \\
9                     & IITK               & 0.45             \\
10                    & UCSC NLP           & 0.45             \\
11                    & CLTeam1            & 0.44             \\
13                    & UMUTeam            & 0.43             \\
12                    & ISDS-NLP           & 0.43             \\
14                    & BITS Pilani        & 0.42             \\
15                    & AIMA               & 0.42             \\
16                    & SSN\_Semeval10     & 0.40             \\
17                    & Hidetsune          & 0.39             \\
18                    & INGEOTEC           & 0.39             \\
19                    & SSN\_ARMM          & 0.38             \\
20                    & TransMistral       & 0.36             \\
23                    & TECHSSN            & 0.34             \\
24                    & MorphingMinds      & 0.33             \\
26                    & RACAI              & 0.31             \\
27                    & Innovators         & 0.28             \\
30                    & VerbaNexAI Lab     & 0.24             \\
32                    & YNU-HPCC           & 0.18             \\ \hline
\end{tabular}%
}
\caption{Results (Weighted F1) for Task A. Rank is as mentioned in CodaLab. Team Name is as mentioned in the corresponding system description.}
\label{tab:erc_results}
\vspace{-4mm}
\end{table}

\subsection{Task A:  ERC in Hindi-English code-mixed conversation}
The results for Task A are compiled in Table \ref{tab:erc_results}. Out of the $24$ submitted papers, $22$ teams explored the code-mixed ERC task, attaining weighted F1 scores spanning from $0.70$ to $0.18$. Notably, the foremost twelve teams, up to SSN\_Semeval10, opted for LLMs as their architectural preference, yielding top performances. Following closely, RNN-based approaches such as LSTM and classical ML methods like SVM emerged as the subsequent choices. A notable observation is the substantial disparity (approximately $37\%$) in performance between the leading model and the succeeding system.

Both leading teams relied on LLMs as their primary architectural framework, yet IASBS diverged by integrating classical ML methods. Their innovative two-phase strategy, combining LLMs for contextual representations and ML techniques for classification, evidently yielded significant improvements. Conversely, the subsequent top model utilised LLMs without any ensembling. Team FeedForward, securing fifth place on the CodaLab leaderboard for Task A, implemented instruction-based finetuning and quantized low-rank adaptation alongside novel techniques like sentext-height and enhanced prompting strategies.

Another intriguing observation arises from the marginal discrepancy (approximately $2\%$) between strategies based on LLMs and those employing classical ML techniques. Team SSN\_Semeval10 refined a BERT classifier, achieving a weighted F1-score of $0.40$. Conversely, team Hidetsune took a different approach by translating all code-mixed data into English and employing data augmentation to bolster the 'English'-based ERC dataset. Subsequently, they trained a SpaCy-v3\footnote{\url{https://spacy.io/}} classifier, resulting in a weighted F1-score of $0.39$.

\begin{table}[t]
\centering
\resizebox{\columnwidth}{!}{%
\begin{tabular}{l|p{15em}|l}
\hline
\textbf{Rank} & \textbf{Team Name} & \textbf{Results} \\ \hline
1             & TW-NLP             & 0.79                                  \\
2             & Innovators         & 0.79                                  \\
2             & UCSC NLP           & 0.79                                  \\
2             & GAVx               & 0.79                                  \\
3             & FeedForward        & 0.77                                  \\
5             & IITK               & 0.56                                  \\
6             & UMUTeam            & 0.26                                  \\
7             & IASBS              & 0.12                                  \\
9             & SSN\_ARMM          & 0.11                                  \\
11            & TECHSSN            & 0.1                                   \\
21            & YNU-HPCC           & 0.01                                  \\ \hline
\end{tabular}%
}
\caption{Results for Task B. F1 score for trigger utterances is our metric of choice. Rank is as mentioned in CodaLab. Team Name is as mentioned in the corresponding system description.}
\label{tab:efr_cm_results}
\vspace{-4mm}
\end{table}

\subsection{Task B – EFR in Hindi-English code-mixed conversation}
\label{sec:efr_analysis}
Table \ref{tab:efr_cm_results} presents the outcomes for Task B, wherein the highest performance attained a trigger F1-score of $0.79$. Particularly intriguing is the fact that the leading four teams achieved identical F1-scores, with the top two teams opting for conventional ML and rule-based approaches. This phenomenon stems from the common occurrence wherein a speaker's emotional shift in a conversation at utterance $i$ is predominantly triggered by the $i-1$ utterance. This pattern underscores the significance of the preceding utterance as a trigger. Illustrated in Figure \ref{fig:trigdist} is the trigger distribution within the dialogues of \efrCM\ and \efrEnglish. Evidently, the majority of trigger utterances are the $i-1^{th}$ utterances. Employing XGBoost for trigger classification, the leading team, TW-NLP, secured their position, while the second-ranking team opted for a rule-based approach, designating all $i-1$ utterances as triggers. This strategy led to the attainment of the highest score of $0.79$ F1.

\subsection{Task C – EFR in English conversation}
\begin{figure}[t]
    \centering
    \includegraphics[width=\columnwidth]{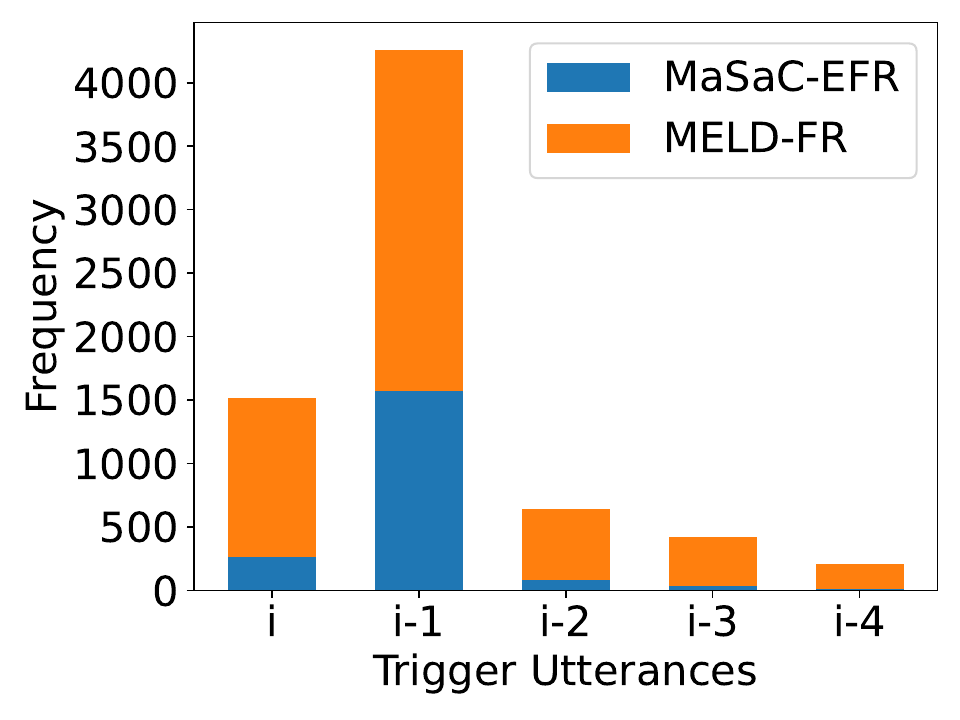}
    \caption{Distribution of triggers for the last four utterances from the trigger utterance $i$.}
    \label{fig:trigdist}
    \vspace{-2mm}
\end{figure}

The outcomes for Task C are displayed in Table \ref{tab:efr_eng_results}, revealing the top-performing system achieving an F1 score of $0.76$ for the triggers. Impressively, the subsequent results closely trail the best one, exhibiting only a marginal gap of approximately $2\% to 4\%$. Notably, the leading two performers in the task predominantly utilise methods employing LLMs, while the third-best performance is attributed to XGBoost. Illustrated in Figure \ref{fig:trigdist}, \efrEnglish\ also grapples with a skewed distribution of trigger utterances, thereby resulting in comparable performances between LLMs and ML-based systems.

\begin{table}[t]
\centering
\resizebox{\columnwidth}{!}{%
\begin{tabular}{l|p{15em}|l}
\hline
\textbf{Rank} & \textbf{Team Name} & \textbf{Results} \\ \hline
2             & GAVx               & 0.76                                  \\
3             & FeedForward        & 0.74                                  \\
5             & TW-NLP             & 0.71                                  \\
7             & Innovators         & 0.68                                  \\
8             & UCSC NLP           & 0.68                                  \\
10            & IITK               & 0.6                                   \\
11            & SSN\_ARMM          & 0.26                                  \\
12            & IASBS              & 0.25                                  \\
13            & TECHSSN            & 0.24                                  \\
15            & UMUTeam            & 0.22                                  \\
26            & YNU-HPCC           & 0.07                                  \\ \hline
\end{tabular}%
}
\caption{Results for Task C. F1 score for trigger utterances is our metric of choice. Rank is as mentioned in CodaLab. Team Name is as mentioned in the corresponding system description.}
\label{tab:efr_eng_results}
\vspace{-5mm}
\end{table}

\subsection{Findings by Participants}
\begin{figure}[t]
    \centering
    \includegraphics[width=\columnwidth]{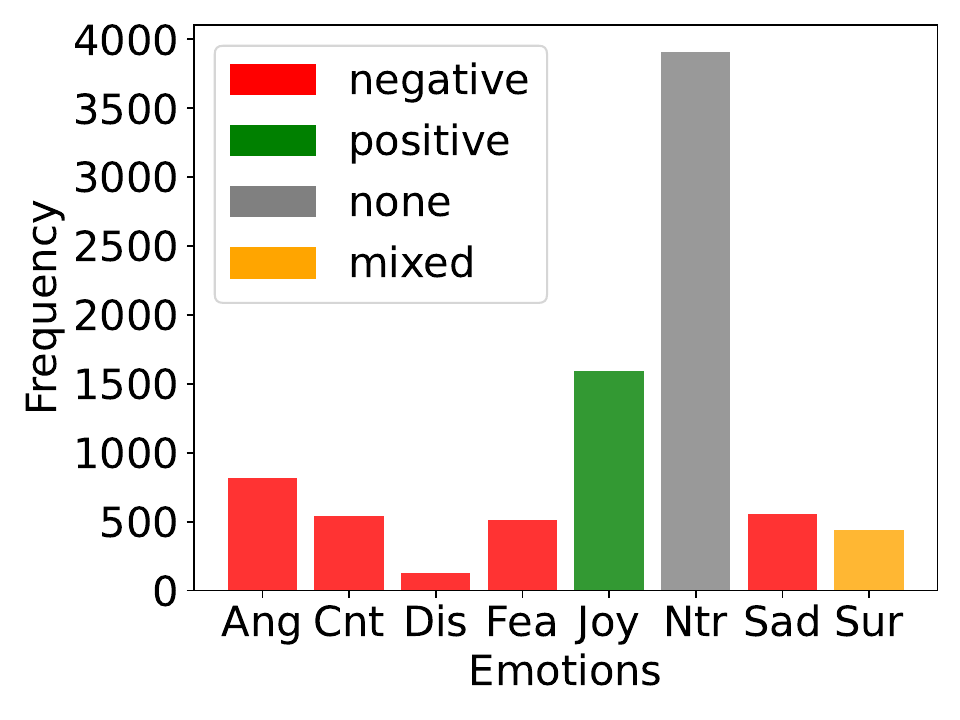}
    \caption{Emotion distribution in \ercCM. The colors depict the distribution of emotions capturing positive, negative, mixed, and no feelings (Abbreviations: Ang: Anger, Cnt: Contempt, Dis: Disgust, Fea: Fear, Ntr: Neutral, Sad: Sadness, Sur: Surprise).}
    \label{fig:emodist}
    \vspace{-4mm}
\end{figure}

\textbf{Challenge of code-mixing.} The dataset utilised in this shared task encompasses Hindi-English code-mixed instances for subtasks A and B, presenting the most formidable challenge of the competition. To address this hurdle, several teams, including TransMistral, FeedForward, and Hidetsune, opted for translation, converting all code-mixed instances into monolingual English before engaging in any classification process. Additionally, teams such as TW-NLP leveraged multilingual LLMs like MBERT to effectively manage code-mixed input.\\
\textbf{Effect of data augmentation.} Machine learning and deep learning techniques exhibit an insatiable appetite for data, giving rise to circumstances where an abundance of data tends to correlate with improved performance. In light of this conjecture, several teams, including Hidetsune, ventured into experimenting with data augmentation for the ERC task. The general observation revealed an enhancement in performance with the incorporation of more data during model training.\\
\textbf{Required context for classification.} Emotions are fleeting and are typically influenced by the immediate circumstances in which the speaker finds themselves. As a result, the nearby utterances within a dialogue exert a more pronounced impact on determining the emotional nuances of a speaker compared to utterances further removed in context. This phenomenon is depicted in Figure \ref{fig:trigdist}. Consequently, teams such as FeedForward and IITK initially ascertain the requisite extent of context needed for conducting ERC, before proceeding with classification, taking the computed context into consideration.\\
\textbf{Challenge of implicit triggers.} Emotion flips can generally be attributed to two scenarios: firstly, when something uttered in the dialogue directly prompts the emotion flip, constituting explicit triggers; and secondly, when events external to the dialogue, such as an act of theft, occur without explicit mention in the dialogue, representing implicit triggers. In both the \efrCM\ and \efrEnglish\ datasets, instances of implicit triggers exist where no trigger utterances are marked in the dialogue. These instances present a challenge for the learned models of several teams, including GAVx.\\
\textbf{Negative vs positive emotions.} The dataset \ercCM\ utilises Ekman emotions \cite{ekman} as its set of emotion labels, encompassing six emotions and one label for neutral emotions. These emotions include Anger, Contempt, Disgust, Fear, Joy, Neutral, Sadness, and Surprise. Notably, among these emotions, five portray negative feelings (Anger, Contempt, Disgust, Fear, and Sadness), while only one represents positive emotions (Joy). Surprise, on the other hand, can convey either positive or negative emotions depending on the context. Figure \ref{fig:emodist} displays the distribution of these emotions within \ercCM. It's evident that as there's only one category for positive emotions, all such instances are classified as joy, leading to a higher frequency of joy compared to other emotions. Moreover, the neutral category has the most instances compared to the others. Consequently, many teams, like IITK, have noted that their models perform better for the neutral and joy labels than for any other emotion.

\begin{table}[h!]
\centering
\resizebox{0.9\columnwidth}{!}{%
\begin{tabular}{l|l|l|l}
\hline
\textbf{Team}  & \textbf{Task A} & \textbf{Task B} & \textbf{Task C} \\ \hline
MasonTigers    & \cellcolor{green!80}0.78 (1)        & \cellcolor{yellow!80}0.79 (2)        & \cellcolor{green!80}0.79 (1)        \\
Knowdee        & \cellcolor{yellow!80}0.73 (2)        & 0.66 (4)        & 0.61 (9)        \\
IASBS          & \cellcolor{orange!80}0.70 (3)        & 0.12 (7)        & 0.25 (12)       \\
-              & 0.66 (4)        & 0.07 (20)       & 0.04 (28)       \\
FeedForward    & 0.51 (5)        & \cellcolor{orange!80}0.77 (3)        & \cellcolor{orange!80}0.74 (3)        \\
TW-NLP         & 0.45 (6)        & \cellcolor{green!80}0.79 (1)        & 0.71 (5)        \\
TechSSN1       & 0.45 (7)        & 0.00 (22)       & 0.00 (29)       \\
-              & 0.45 (8)        & \cellcolor{yellow!80}0.79 (2)        & 0.68 (8)        \\
IITK           & 0.45 (9)        & 0.56 (5)        & 0.60 (10)       \\
-              & 0.45 (10)       & 0.10 (11)       & 0.15 (23)       \\
CLTeam1        & 0.44 (11)       & 0.10 (11)       & 0.24 (13)       \\
UniBucNLP      & 0.43 (12)       & 0.10 (14)       & 0.06 (27)       \\
UMUTeam        & 0.43 (13)       & 0.26 (6)        & 0.22 (15)       \\
BITS Pilani    & 0.42 (14)       & 0.10 (11)       & 0.24 (13)       \\
AIMA           & 0.42 (15)       & 0.10 (12)       & 0.21 (21)       \\
SSN\_Semeval10 & 0.40 (16)       & 0.00 (22)       & 0.00 (29)       \\
OZemi          & 0.39 (17)       & 0.00 (22)       & 0.00 (29)       \\
INGEOTEC       & 0.39 (18)       & 0.00 (22)       & 0.00 (29)       \\
-              & 0.38 (19)       & 0.11 (9)        & 0.26 (11)       \\
-              & 0.37 (20)       & 0.07 (19)       & 0.07 (25)       \\
CUET\_NLP      & 0.37 (21)       & 0.00 (22)       & 0.00 (29)       \\
-              & 0.36 (22)       & 0.10 (10)       & 0.22 (19)       \\
TechSSN        & 0.34 (23)       & 0.10 (11)       & 0.24 (13)       \\
MorphingMinds  & 0.33 (24)       & 0.10 (16)       & 0.22 (18)       \\
Z-AGI Labs     & 0.31 (25)       & 0.00 (22)       & 0.00 (29)       \\
RACAI          & 0.31 (26)       & 0.10 (11)       & 0.24 (13)       \\
Innovators     & 0.28 (27)       & \cellcolor{yellow!80}0.79 (2)        & 0.68 (7)        \\
-              & 0.27 (28)       & 0.10 (13)       & 0.21 (20)       \\
-              & 0.26 (29)       & 0.10 (15)       & 0.16 (22)       \\
-              & 0.24 (30)       & 0.00 (22)       & 0.74 (4)        \\
LinguisTech    & 0.24 (30)       & 0.00 (22)       & 0.70 (6)        \\
Team + 1       & 0.24 (30)       & 0.09 (17)       & 0.22 (16)       \\
PartOfGlitch   & 0.24 (30)       & 0.00 (22)       & 0.10 (24)       \\
VerbNexAI Lab  & 0.24 (30)       & 0.00 (22)       & 0.00 (29)       \\
-              & 0.24 (30)       & 0.00 (22)       & 0.00 (29)       \\
silp\_nlp      & 0.24 (31)       & 0.11 (8)        & 0.23 (14)       \\
-              & 0.18 (32)       & 0.01 (21)       & 0.07 (26)       \\
-              & 0.14 (33)       & 0.09 (18)       & 0.22 (17)       \\
GAVx           & 0.08 (34)       & \cellcolor{yellow!80}0.79 (2)        & \cellcolor{yellow!80}0.76 (2)        \\ \hline
\end{tabular}%
}
\caption{Leaderboard from CodaLab. Rank for each task is mentioned in parenthesis. Top three systems are highlighted in \colorbox{green!80}{green}, \colorbox{yellow!80}{yellow}, and \colorbox{orange!80}{orange}.}
\label{tab:leaderboard}
\end{table}

\subsection{Leaderboard}
In this paper, we have exclusively examined the outcomes of participants who provided a description of their system(s) for the shared task. The complete array of ranks, team names, and results of the participants in the CodaLab competition, encompassing users who didn't submit a system description, is depicted in Table \ref{tab:leaderboard}. Although not all teams attempted all three tasks, however, they obtained some score in the leaderboard since they submitted some values for the output. A glimpse at the leaderboard reveals that the best performance for tasks A, B, and C stood at $0.78$, $0.79$, and $0.79$, respectively. One, four, and one team(s) achieved best performance of the three tasks.

\section{Conclusion}
\label{sec:conclusion}
This paper outlines SemEval 2024 Task 10, covering its goals, data, participants, and results. It includes three subtasks: emotion identification in code-mixed dialogues and pinpointing triggers for emotion shifts in code-mixed and English dialogues. $84$ participants competed on CodaLab, with $24$ teams submitting system description papers. Top systems for Tasks A and C used LLM-based architectures, while Task B favored standard ML techniques. Leading systems achieved F1 scores of $0.70$, $0.79$, and $0.76$ across subtasks, indicating impressive performance but also highlighting ongoing challenges for future research.

\bibliography{custom}

\begin{thebibliography}{50}
\expandafter\ifx\csname natexlab\endcsname\relax\def\natexlab#1{#1}\fi

\bibitem[{Bedi et~al.(2023)Bedi, Kumar, Akhtar, and Chakraborty}]{10.1109/TAFFC.2021.3083522}
Manjot Bedi, Shivani Kumar, Md~Shad Akhtar, and Tanmoy Chakraborty. 2023.
\newblock \href {https://doi.org/10.1109/TAFFC.2021.3083522} {Multi-modal sarcasm detection and humor classification in code-mixed conversations}.
\newblock \emph{IEEE Trans. Affect. Comput.}, 14(2):1363–1375.

\bibitem[{Bosselut et~al.(2019)Bosselut, Rashkin, Sap, Malaviya, Celikyilmaz, and Choi}]{bosselut-etal-2019-comet}
Antoine Bosselut, Hannah Rashkin, Maarten Sap, Chaitanya Malaviya, Asli Celikyilmaz, and Yejin Choi. 2019.
\newblock \href {https://doi.org/10.18653/v1/P19-1470} {{COMET}: Commonsense transformers for automatic knowledge graph construction}.
\newblock In \emph{Proceedings of the 57th Annual Meeting of the Association for Computational Linguistics}, pages 4762--4779, Florence, Italy. Association for Computational Linguistics.

\bibitem[{Brown et~al.(2020)Brown, Mann, Ryder, Subbiah, Kaplan, Dhariwal, Neelakantan, Shyam, Sastry, Askell, Agarwal, Herbert-Voss, Krueger, Henighan, Child, Ramesh, Ziegler, Wu, Winter, Hesse, Chen, Sigler, Litwin, Gray, Chess, Clark, Berner, McCandlish, Radford, Sutskever, and Amodei}]{NEURIPS2020_1457c0d6}
Tom Brown, Benjamin Mann, Nick Ryder, Melanie Subbiah, Jared~D Kaplan, Prafulla Dhariwal, Arvind Neelakantan, Pranav Shyam, Girish Sastry, Amanda Askell, Sandhini Agarwal, Ariel Herbert-Voss, Gretchen Krueger, Tom Henighan, Rewon Child, Aditya Ramesh, Daniel Ziegler, Jeffrey Wu, Clemens Winter, Chris Hesse, Mark Chen, Eric Sigler, Mateusz Litwin, Scott Gray, Benjamin Chess, Jack Clark, Christopher Berner, Sam McCandlish, Alec Radford, Ilya Sutskever, and Dario Amodei. 2020.
\newblock \href {https://proceedings.neurips.cc/paper_files/paper/2020/file/1457c0d6bfcb4967418bfb8ac142f64a-Paper.pdf} {Language models are few-shot learners}.
\newblock In \emph{Advances in Neural Information Processing Systems}, volume~33, pages 1877--1901. Curran Associates, Inc.

\bibitem[{Chuang and Wu(2004)}]{chuang2004multi}
Ze-Jing Chuang and Chung-Hsien Wu. 2004.
\newblock Multi-modal emotion recognition from speech and text.
\newblock In \emph{International Journal of Computational Linguistics \& Chinese Language Processing, Volume 9, Number 2, August 2004: Special Issue on New Trends of Speech and Language Processing}, pages 45--62.

\bibitem[{Cowen and Keltner(2017)}]{cowen2017self}
Alan~S Cowen and Dacher Keltner. 2017.
\newblock Self-report captures 27 distinct categories of emotion bridged by continuous gradients.
\newblock \emph{PNAS}, 114(38):E7900--E7909.

\bibitem[{Cui et~al.(2020)Cui, Liu, Zhang, Chen, Wang, and Chen}]{cui2020eeg}
Heng Cui, Aiping Liu, Xu~Zhang, Xiang Chen, Kongqiao Wang, and Xun Chen. 2020.
\newblock Eeg-based emotion recognition using an end-to-end regional-asymmetric convolutional neural network.
\newblock \emph{Knowledge-Based Systems}, 205:106243.

\bibitem[{Devlin et~al.(2018)Devlin, Chang, Lee, and Toutanova}]{devlin2018bert}
Jacob Devlin, Ming-Wei Chang, Kenton Lee, and Kristina Toutanova. 2018.
\newblock Bert: Pre-training of deep bidirectional transformers for language understanding.
\newblock \emph{arXiv preprint arXiv:1810.04805}.

\bibitem[{Ekman(1992)}]{ekman}
Paul Ekman. 1992.
\newblock An argument for basic emotions.
\newblock \emph{Cognition \& emotion}, 6(3-4):169--200.

\bibitem[{Fitrianie et~al.(2003)Fitrianie, Wiggers, and Rothkrantz}]{fitrianie2003multi}
Siska Fitrianie, Pascal Wiggers, and Leon~JM Rothkrantz. 2003.
\newblock A multi-modal eliza using natural language processing and emotion recognition.
\newblock In \emph{Text, Speech and Dialogue: 6th International Conference, TSD 2003, {\v{C}}esk{\'e} Bud{\'e}jovice, Czech Republic, September 8-12, 2003. Proceedings 6}, pages 394--399. Springer.

\bibitem[{Ghosal et~al.(2019)Ghosal, Majumder, Poria, Chhaya, and Gelbukh}]{ghosal-etal-2019-dialoguegcn}
Deepanway Ghosal, Navonil Majumder, Soujanya Poria, Niyati Chhaya, and Alexander Gelbukh. 2019.
\newblock {D}ialogue{GCN}: A graph convolutional neural network for emotion recognition in conversation.
\newblock In \emph{Proceedings of the 2019 Conference on Empirical Methods in Natural Language Processing and the 9th International Joint Conference on Natural Language Processing (EMNLP-IJCNLP)}, pages 154--164, Hong Kong, China.

\bibitem[{Hazarika et~al.(2018{\natexlab{a}})Hazarika, Poria, Mihalcea, Cambria, and Zimmermann}]{hazarika-etal-2018-icon}
Devamanyu Hazarika, Soujanya Poria, Rada Mihalcea, Erik Cambria, and Roger Zimmermann. 2018{\natexlab{a}}.
\newblock {ICON}: Interactive conversational memory network for multimodal emotion detection.
\newblock In \emph{Proceedings of the 2018 Conference on Empirical Methods in Natural Language Processing}, pages 2594--2604, Brussels, Belgium.

\bibitem[{Hazarika et~al.(2018{\natexlab{b}})Hazarika, Poria, Zadeh, Cambria, Morency, and Zimmermann}]{hazarika-etal-2018-conversational}
Devamanyu Hazarika, Soujanya Poria, Amir Zadeh, Erik Cambria, Louis-Philippe Morency, and Roger Zimmermann. 2018{\natexlab{b}}.
\newblock Conversational memory network for emotion recognition in dyadic dialogue videos.
\newblock In \emph{Proceedings of the 2018 Conference of the North {A}merican Chapter of the Association for Computational Linguistics: Human Language Technologies, Volume 1 (Long Papers)}, pages 2122--2132, New Orleans, Louisiana.

\bibitem[{Hazarika et~al.(2021)Hazarika, Poria, Zimmermann, and Mihalcea}]{tlerc}
Devamanyu Hazarika, Soujanya Poria, Roger Zimmermann, and Rada Mihalcea. 2021.
\newblock Conversational transfer learning for emotion recognition.
\newblock \emph{Information Fusion}, 65:1--12.

\bibitem[{Ilyas et~al.(2023)Ilyas, Shahzad, and Kamran~Malik}]{10.1145/3552515}
Abdullah Ilyas, Khurram Shahzad, and Muhammad Kamran~Malik. 2023.
\newblock \href {https://doi.org/10.1145/3552515} {Emotion detection in code-mixed roman urdu - english text}.
\newblock \emph{ACM Trans. Asian Low-Resour. Lang. Inf. Process.}, 22(2).

\bibitem[{Jiang et~al.(2023)Jiang, Sablayrolles, Mensch, Bamford, Chaplot, de~las Casas, Bressand, Lengyel, Lample, Saulnier, Lavaud, Lachaux, Stock, Scao, Lavril, Wang, Lacroix, and Sayed}]{jiang2023mistral}
Albert~Q. Jiang, Alexandre Sablayrolles, Arthur Mensch, Chris Bamford, Devendra~Singh Chaplot, Diego de~las Casas, Florian Bressand, Gianna Lengyel, Guillaume Lample, Lucile Saulnier, Lélio~Renard Lavaud, Marie-Anne Lachaux, Pierre Stock, Teven~Le Scao, Thibaut Lavril, Thomas Wang, Timothée Lacroix, and William~El Sayed. 2023.
\newblock \href {http://arxiv.org/abs/2310.06825} {Mistral 7b}.

\bibitem[{Jiao et~al.(2020)Jiao, Lyu, and King}]{aghmn}
Wenxiang Jiao, Michael Lyu, and Irwin King. 2020.
\newblock Real-time emotion recognition via attention gated hierarchical memory network.
\newblock In \emph{AAAI}, volume~34, pages 8002--8009.

\bibitem[{Jiao et~al.(2019)Jiao, Yang, King, and Lyu}]{jiao2019higru}
Wenxiang Jiao, Haiqin Yang, Irwin King, and Michael~R Lyu. 2019.
\newblock Higru: Hierarchical gated recurrent units for utterance-level emotion recognition.
\newblock \emph{arXiv preprint arXiv:1904.04446}.

\bibitem[{Krippendorff(2011)}]{krippendorff2011computing}
Klaus Krippendorff. 2011.
\newblock Computing krippendorff's alpha-reliability.

\bibitem[{Kumar et~al.(2023{\natexlab{a}})Kumar, Dudeja, Akhtar, and Chakraborty}]{10164178}
Shivani Kumar, Shubham Dudeja, Md~Shad Akhtar, and Tanmoy Chakraborty. 2023{\natexlab{a}}.
\newblock \href {https://doi.org/10.1109/TAI.2023.3289937} {Emotion flip reasoning in multiparty conversations}.
\newblock \emph{IEEE Transactions on Artificial Intelligence}, pages 1--10.

\bibitem[{Kumar et~al.(2022{\natexlab{a}})Kumar, Kulkarni, Akhtar, and Chakraborty}]{kumar-etal-2022-become}
Shivani Kumar, Atharva Kulkarni, Md~Shad Akhtar, and Tanmoy Chakraborty. 2022{\natexlab{a}}.
\newblock \href {https://doi.org/10.18653/v1/2022.acl-long.411} {When did you become so smart, oh wise one?! sarcasm explanation in multi-modal multi-party dialogues}.
\newblock In \emph{Proceedings of the 60th Annual Meeting of the Association for Computational Linguistics (Volume 1: Long Papers)}, pages 5956--5968, Dublin, Ireland. Association for Computational Linguistics.

\bibitem[{Kumar et~al.(2022{\natexlab{b}})Kumar, Mondal, Akhtar, and Chakraborty}]{kumar2022explaining}
Shivani Kumar, Ishani Mondal, Md~Shad Akhtar, and Tanmoy Chakraborty. 2022{\natexlab{b}}.
\newblock \href {http://arxiv.org/abs/2211.11049} {Explaining (sarcastic) utterances to enhance affect understanding in multimodal dialogues}.

\bibitem[{Kumar et~al.(2023{\natexlab{b}})Kumar, S, Akhtar, and Chakraborty}]{kumar-etal-2023-multilingual}
Shivani Kumar, Ramaneswaran S, Md~Akhtar, and Tanmoy Chakraborty. 2023{\natexlab{b}}.
\newblock \href {https://doi.org/10.18653/v1/2023.emnlp-main.598} {From multilingual complexity to emotional clarity: Leveraging commonsense to unveil emotions in code-mixed dialogues}.
\newblock In \emph{Proceedings of the 2023 Conference on Empirical Methods in Natural Language Processing}, pages 9638--9652, Singapore. Association for Computational Linguistics.

\bibitem[{Kumar et~al.(2021)Kumar, Shrimal, Akhtar, and Chakraborty}]{kumar2021discovering}
Shivani Kumar, Anubhav Shrimal, Md~Shad Akhtar, and Tanmoy Chakraborty. 2021.
\newblock Discovering emotion and reasoning its flip in multi-party conversations using masked memory network and transformer.
\newblock \emph{arXiv 2103.12360 (cs.CL)}.

\bibitem[{Lee et~al.(2010)Lee, Chen, and Huang}]{lee-etal-2010-text}
Sophia Yat~Mei Lee, Ying Chen, and Chu-Ren Huang. 2010.
\newblock A text-driven rule-based system for emotion cause detection.
\newblock In \emph{Proceedings of the {NAACL} {HLT} 2010 Workshop on Computational Approaches to Analysis and Generation of Emotion in Text}, pages 45--53, Los Angeles, CA.

\bibitem[{Li et~al.(2007)Li, Pang, Guo, and Wang}]{li2007research}
Haifang Li, Na~Pang, Shangbo Guo, and Heping Wang. 2007.
\newblock Research on textual emotion recognition incorporating personality factor.
\newblock In \emph{2007 IEEE International Conference on Robotics and Biomimetics (ROBIO)}, pages 2222--2227. IEEE.

\bibitem[{Li et~al.(2020)Li, Shao, Ji, and Cambria}]{bieru}
Wei Li, Wei Shao, Shaoxiong Ji, and Erik Cambria. 2020.
\newblock Bieru: bidirectional emotional recurrent unit for conversational sentiment analysis.
\newblock \emph{arXiv preprint arXiv:2006.00492}.

\bibitem[{Liu et~al.(2019)Liu, Ott, Goyal, Du, Joshi, Chen, Levy, Lewis, Zettlemoyer, and Stoyanov}]{liu2019roberta}
Yinhan Liu, Myle Ott, Naman Goyal, Jingfei Du, Mandar Joshi, Danqi Chen, Omer Levy, Mike Lewis, Luke Zettlemoyer, and Veselin Stoyanov. 2019.
\newblock Roberta: A robustly optimized bert pretraining approach.
\newblock \emph{arXiv preprint arXiv:1907.11692}.

\bibitem[{Ma et~al.(2022)Ma, Wang, Lin, Pan, Zhang, and Yang}]{ma2022multi}
Hui Ma, Jian Wang, Hongfei Lin, Xuejun Pan, Yijia Zhang, and Zhihao Yang. 2022.
\newblock A multi-view network for real-time emotion recognition in conversations.
\newblock \emph{Knowledge-Based Systems}, 236:107751.

\bibitem[{Madhu et~al.(2023)Madhu, Satapara, Modha, Mandl, and Majumder}]{madhu2023detecting}
Hiren Madhu, Shrey Satapara, Sandip Modha, Thomas Mandl, and Prasenjit Majumder. 2023.
\newblock Detecting offensive speech in conversational code-mixed dialogue on social media: A contextual dataset and benchmark experiments.
\newblock \emph{Expert Systems with Applications}, 215:119342.

\bibitem[{Mencattini et~al.(2014)Mencattini, Martinelli, Costantini, Todisco, Basile, Bozzali, and Di~Natale}]{mencattini2014speech}
Arianna Mencattini, Eugenio Martinelli, Giovanni Costantini, Massimiliano Todisco, Barbara Basile, Marco Bozzali, and Corrado Di~Natale. 2014.
\newblock Speech emotion recognition using amplitude modulation parameters and a combined feature selection procedure.
\newblock \emph{Knowledge-Based Systems}, 63:68--81.

\bibitem[{OpenAI et~al.(2023)OpenAI, :, and et~al.}]{openai2023gpt4}
OpenAI, :, and Josh~Achiam et~al. 2023.
\newblock \href {http://arxiv.org/abs/2303.08774} {Gpt-4 technical report}.

\bibitem[{Picard(1997)}]{Picard:1997:AC:265013}
Rosalind~W. Picard. 1997.
\newblock \emph{{Affective Computing}}.
\newblock MIT Press, Cambridge, MA, USA.

\bibitem[{Poria et~al.(2017)Poria, Cambria, Hazarika, Majumder, Zadeh, and Morency}]{poria2017context}
Soujanya Poria, Erik Cambria, Devamanyu Hazarika, Navonil Majumder, Amir Zadeh, and Louis-Philippe Morency. 2017.
\newblock Context-dependent sentiment analysis in user-generated videos.
\newblock In \emph{ACL}, pages 873--883.

\bibitem[{Poria et~al.(2019)Poria, Hazarika, Majumder, Naik, Cambria, and Mihalcea}]{poria-etal-2019-meld}
Soujanya Poria, Devamanyu Hazarika, Navonil Majumder, Gautam Naik, Erik Cambria, and Rada Mihalcea. 2019.
\newblock {MELD}: A multimodal multi-party dataset for emotion recognition in conversations.
\newblock In \emph{Proceedings of the 57th Annual Meeting of the Association for Computational Linguistics}, pages 527--536, Florence, Italy.

\bibitem[{Poria et~al.(2021)Poria, Majumder, Hazarika, Ghosal, Bhardwaj, Jian, Hong, Ghosh, Roy, Chhaya et~al.}]{poria2021recognizing}
Soujanya Poria, Navonil Majumder, Devamanyu Hazarika, Deepanway Ghosal, Rishabh Bhardwaj, Samson Yu~Bai Jian, Pengfei Hong, Romila Ghosh, Abhinaba Roy, Niyati Chhaya, et~al. 2021.
\newblock Recognizing emotion cause in conversations.
\newblock \emph{Cognitive Computation}, 13:1317--1332.

\bibitem[{Radford et~al.(2019)Radford, Wu, Child, Luan, Amodei, Sutskever et~al.}]{radford2019language}
Alec Radford, Jeffrey Wu, Rewon Child, David Luan, Dario Amodei, Ilya Sutskever, et~al. 2019.
\newblock Language models are unsupervised multitask learners.
\newblock \emph{OpenAI blog}, 1(8):9.

\bibitem[{Sanh et~al.(2020)Sanh, Debut, Chaumond, and Wolf}]{sanh2020distilbert}
Victor Sanh, Lysandre Debut, Julien Chaumond, and Thomas Wolf. 2020.
\newblock \href {http://arxiv.org/abs/1910.01108} {Distilbert, a distilled version of bert: smaller, faster, cheaper and lighter}.

\bibitem[{Sasidhar et~al.(2020)Sasidhar, B, and P}]{SASIDHAR20201346}
T~Tulasi Sasidhar, Premjith B, and Soman~K P. 2020.
\newblock \href {https://doi.org/https://doi.org/10.1016/j.procs.2020.04.144} {Emotion detection in hinglish(hindi+english) code-mixed social media text}.
\newblock \emph{Procedia Computer Science}, 171:1346--1352.
\newblock Third International Conference on Computing and Network Communications (CoCoNet'19).

\bibitem[{Shen et~al.(2020)Shen, Chen, Quan, and Xie}]{dialogxl}
Weizhou Shen, Junqing Chen, Xiaojun Quan, and Zhixian Xie. 2020.
\newblock Dialogxl: All-in-one xlnet for multi-party conversation emotion recognition.
\newblock \emph{arXiv preprint arXiv:2012.08695}.

\bibitem[{Suciati and Budi(2020)}]{Suciati2020}
Andi Suciati and Indra Budi. 2020.
\newblock \href {https://doi.org/10.14569/IJACSA.2020.0110921} {Aspect-based sentiment analysis and emotion detection for code-mixed review}.
\newblock \emph{International Journal of Advanced Computer Science and Applications}, 11(9).

\bibitem[{Touvron et~al.(2023)Touvron, Lavril, Izacard, Martinet, Lachaux, Lacroix, Rozière, Goyal, Hambro, Azhar, Rodriguez, Joulin, Grave, and Lample}]{touvron2023llama}
Hugo Touvron, Thibaut Lavril, Gautier Izacard, Xavier Martinet, Marie-Anne Lachaux, Timothée Lacroix, Baptiste Rozière, Naman Goyal, Eric Hambro, Faisal Azhar, Aurelien Rodriguez, Armand Joulin, Edouard Grave, and Guillaume Lample. 2023.
\newblock \href {http://arxiv.org/abs/2302.13971} {Llama: Open and efficient foundation language models}.

\bibitem[{Tu et~al.(2022)Tu, Wen, Liu, Jiang, and Cambria}]{9706271}
Geng Tu, Jintao Wen, Cheng Liu, Dazhi Jiang, and Erik Cambria. 2022.
\newblock \href {https://doi.org/10.1109/TAI.2022.3149234} {Context- and sentiment-aware networks for emotion recognition in conversation}.
\newblock \emph{IEEE Transactions on Artificial Intelligence}, 3(5):699--708.

\bibitem[{Tunstall et~al.(2023)Tunstall, Beeching, Lambert, Rajani, Rasul, Belkada, Huang, von Werra, Fourrier, Habib, Sarrazin, Sanseviero, Rush, and Wolf}]{tunstall2023zephyr}
Lewis Tunstall, Edward Beeching, Nathan Lambert, Nazneen Rajani, Kashif Rasul, Younes Belkada, Shengyi Huang, Leandro von Werra, Clémentine Fourrier, Nathan Habib, Nathan Sarrazin, Omar Sanseviero, Alexander~M. Rush, and Thomas Wolf. 2023.
\newblock \href {http://arxiv.org/abs/2310.16944} {Zephyr: Direct distillation of lm alignment}.

\bibitem[{Wadhawan and Aggarwal(2021)}]{wadhawan2021emotion}
Anshul Wadhawan and Akshita Aggarwal. 2021.
\newblock \href {http://arxiv.org/abs/2102.09943} {Towards emotion recognition in hindi-english code-mixed data: A transformer based approach}.

\bibitem[{Wang et~al.(2023)Wang, Ding, Xia, Li, and Yu}]{9969873}
Fanfan Wang, Zixiang Ding, Rui Xia, Zhaoyu Li, and Jianfei Yu. 2023.
\newblock \href {https://doi.org/10.1109/TAFFC.2022.3226559} {Multimodal emotion-cause pair extraction in conversations}.
\newblock \emph{IEEE Transactions on Affective Computing}, 14(3):1832--1844.

\bibitem[{Yang et~al.(2019)Yang, Li, Wong, Chao, Wang, and Tu}]{Yang_Li_Wong_Chao_Wang_Tu_2019}
Baosong Yang, Jian Li, Derek~F. Wong, Lidia~S. Chao, Xing Wang, and Zhaopeng Tu. 2019.
\newblock \href {https://doi.org/10.1609/aaai.v33i01.3301387} {Context-aware self-attention networks}.
\newblock \emph{Proceedings of the AAAI Conference on Artificial Intelligence}, 33(01):387--394.

\bibitem[{Yang et~al.(2022)Yang, Shen, Mao, and Cai}]{yang2022hybrid}
Lin Yang, Yi~Shen, Yue Mao, and Longjun Cai. 2022.
\newblock Hybrid curriculum learning for emotion recognition in conversation.
\newblock In \emph{Proceedings of the AAAI Conference on Artificial Intelligence}, volume~36, pages 11595--11603.

\bibitem[{Zhang et~al.(2016)Zhang, Mistry, Neoh, and Lim}]{zhang2016intelligent}
Li~Zhang, Kamlesh Mistry, Siew~Chin Neoh, and Chee~Peng Lim. 2016.
\newblock Intelligent facial emotion recognition using moth-firefly optimization.
\newblock \emph{Knowledge-Based Systems}, 111:248--267.

\bibitem[{Zhong et~al.(2019)Zhong, Wang, and Miao}]{zhong-etal-2019-knowledge}
Peixiang Zhong, Di~Wang, and Chunyan Miao. 2019.
\newblock Knowledge-enriched transformer for emotion detection in textual conversations.
\newblock In \emph{Proceedings of the 2019 Conference on Empirical Methods in Natural Language Processing and the 9th International Joint Conference on Natural Language Processing (EMNLP-IJCNLP)}, pages 165--176, Hong Kong, China.

\bibitem[{Zhu et~al.(2022)Zhu, Lou, Deng, and Ji}]{ZHU2022107436}
Xun Zhu, Yinxia Lou, Hongtao Deng, and Donghong Ji. 2022.
\newblock \href {https://doi.org/https://doi.org/10.1016/j.knosys.2021.107436} {Leveraging bilingual-view parallel translation for code-switched emotion detection with adversarial dual-channel encoder}.
\newblock \emph{Knowledge-Based Systems}, 235:107436.

\end{thebibliography}

\end{document}